\documentclass[letterpaper, 10 pt, conference]{ieeeconf}  

\IEEEoverridecommandlockouts                              

\usepackage{graphics} 

\usepackage{amsmath}
\usepackage{amssymb}
\usepackage{multicol}
\usepackage{graphicx}
\usepackage{layouts}
\usepackage{wrapfig}
\usepackage{caption}
\usepackage{graphicx,subcaption}
\usepackage{easyReview}
\usepackage{printlen}
\usepackage{bm}
\usepackage{siunitx}
\usepackage{colortbl}	
\usepackage[utf8]{inputenc}
\usepackage{pgf}
\usepackage{tikz}
\usepackage{tikzscale}
\usepackage{pgfplots}
\usepackage[hidelinks]{hyperref}

\usepackage{pgf}
\DeclareMathOperator{\Diag}{Diag}
\pgfplotsset{compat=1.14}
\usepgfplotslibrary{external}
\usepgfplotslibrary[external]
\tikzset{external/force remake}

\newtheorem{remark}{\textbf{Remark}}

\newcommand{\gencoor}{\boldsymbol{q}}
\newcommand{\genvel}{\boldsymbol{v}}
\newcommand{\nonlinear}{\boldsymbol{n}}
\newcommand{\torque}{\boldsymbol{\tau}}
\newcommand{\force}{\boldsymbol{\lambda}}
\newcommand{\centroidal}{\boldsymbol{h}_G}
\newcommand{\centroidalrate}{\dot{\boldsymbol{h}}_{G}}
\newcommand{\lmom}{\boldsymbol{l}_G}
\newcommand{\amom}{\boldsymbol{k}_G}
\newcommand{\com}{\boldsymbol{c}_G}
\newcommand{\lmomrate}{\dot{\boldsymbol{l}}_G}
\newcommand{\amomrate}{\dot{\boldsymbol{k}}_G}
\newcommand{\comrate}{\dot{\boldsymbol{c}}_G}

\title{\LARGE \bf
On the Use of Torque Measurement in Centroidal State Estimation}

\author{
Shahram Khorshidi$^{1,5}$, Ahmad Gazar$^{1}$, Nicholas Rotella$^{2}$,
Maximilien Naveau$^{3}$,\\ Ludovic Righetti$^{1,4}$, Maren Bennewitz$^{5}$, Majid Khadiv$^{1}$
\thanks{$^{1}$Max-Planck Institute  for  Intelligent  Systems,  T\"ubingen,  Germany. {\tt\small \{firstname.lastname\}@tuebingen.mpg.de}}
\thanks{$^{2}$Agility Robotics, Oregon, USA.
{\tt\small nicholas.rotella@gmail.com}}
\thanks{$^{3}$LAAS-CNRS, Toulouse.
{\tt\small mnaveau@laas.fr}}
\thanks{$^{4}$Tandon  School  of  Engineering,  New  York  University,  USA. {\tt\small ludovic.righetti@nyu.edu}}
\thanks{$^{5}$Humanoid Robots Lab,  University of Bonn,  Germany. {\tt\small maren@cs.uni-bonn.de}, \tt\small{khorshidi@cs.uni-bonn.de}}
\thanks{This work was supported by the Max-Planck Institute for Intelligent Systems' Grassroots program (M10338 and M10343) and the National Science Foundation (CMMI-1825993).}%
}

\begin{document}

\maketitle
\thispagestyle{empty}
\pagestyle{empty}

\begin{abstract}
State-of-the-art legged robots are either capable of measuring torque at the output of their drive systems, or have transparent drive systems which enable the computation of joint torques from motor currents. In either case, this sensor modality is seldom used in state estimation. In this paper, we propose to use joint torque measurements to estimate the centroidal states of legged robots. To do so, we project the whole-body dynamics of a legged robot into the nullspace of the contact constraints, allowing expression of the dynamics independent of the contact forces. Using the constrained dynamics and the centroidal momentum matrix, we are able to directly relate joint torques and centroidal states dynamics. Using the resulting model as the process model of an Extended Kalman Filter (EKF), we fuse the torque measurement in the centroidal state estimation problem. Through real-world experiments on a  quadruped robot executing different gaits, we demonstrate that the estimated centroidal states from our torque-based EKF drastically improve the recovery of these quantities compared to direct computation.
\end{abstract}

\section{Introduction}
State estimation for legged robots plays a crucial role in the successful application of modern state feedback controllers to this domain. The estimation problem is especially difficult for legged robots, as they are inherently underactuated and experience uncertain, intermittent contacts with the environment during motion. Furthermore, their dynamics are highly nonlinear, which makes the design of estimators nontrivial. 

The majority of legged robot estimation works to date have focused on base state estimation. This problem is particularly difficult for floating base systems, as the base pose cannot be measured directly. One of the most widely used choices for base state estimation is the Extended Kalman Filter (EKF) \cite{bloesch2013state,rotella2014state,fallon2014drift}. While there have been recent attempts to use more advanced approaches based on the use of factor graphs \cite{fourmy2021contact} and invariant Kalman Filters \cite{hartley2020contact, ramadoss2021}, the EKF framework is commonly used due to its compromise between simplicity, efficiency, and performance \cite{camurri2020pronto}, \cite{zhang2021}.

For base state estimation, most works have fused Inertial Measurement Unit (IMU) data with leg odometry to estimate the base position and velocity as well as its orientation \cite{bloesch2013state,rotella2014state}. In order to compensate for the drift of the unobservable base position while simultaneously mapping the environment, sensor modalities like LiDAR and cameras are often added \cite{fallon2014drift,camurri2020pronto, stylianos2019}.

Force measurement at the endeffectors of legged robots is another useful source of information that can be used to better estimate \cite{rotella2018contact, geoff2020} and handle contact switch events \cite{bledt2018contact,valsecchi2020quadrupedal} as well as aid in estimation of the centroidal states \cite{rotella2015humanoid,fourmy2021contact}. However, since it is highly important for legged robots to have low leg inertia for agile movements, using sensorized feet \cite{valsecchi2020quadrupedal} can drastically degrade the range of dynamic movements they can perform, especially for quadrupeds. Thus, most of works in this area first estimate the contact forces without direct sensing, and then use the estimated forces for state estimation \cite{bledt2018contact,fourmy2021contact}.

While legged robots often lack endeffector force sensing, they are commonly equipped with joint torque sensors. In this work, we propose to fuse joint torque measurements with the nonlinear dynamic model in the EKF framework to estimate the centroidal states -- namely, the center of mass position and momenta -- of a legged robot. We argue that since most state-of-the-art quadrupedal \cite{hutter2016anymal,katz2019mini,grimminger2020open} and bipedal \cite{chignoli2021humanoid,cassie,daneshmand2021variable} platforms have the capability to measure or estimate joint torque, this sensing modality can be used to perform centroidal state estimation regardless of contact force measurement capabilities. Model-based centroidal state computations are usually very noisy due to noise in measured joint velocities, and low-pass filtering can introduce delays on the order of tens of milliseconds; such large delays can destabilize high rate controllers \cite{rotella2015humanoid}. By using joint torque measurements for centroidal estimation, we can avoid this issue without necessitating contact force sensing. To the best of our knowledge, this is the first time that this sensing modality is used to estimate the centroidal states of a legged robot. Notably, the proposed framework is general enough to be used on any legged robot, not just a quadruped.

The main contributions of this work are as follows:
\begin{itemize}
    \item We propose an EKF-based framework which fuses the whole body dynamics of a legged robot projected into the nullspace of the contact constraints with joint torque measurements in order to perform centroidal state estimation. 
    \item We demonstrate through an extensive set of experiments with different gaits on the real quadruped robot Solo12 \cite{grimminger2020open}, that the use of torque measurements provides accurate centroidal state estimation with minimal noise and time delay.  
\end{itemize}
\section{Fundamentals}
\subsection{Notation}
\begin{itemize}
    \item The $\oplus$ operator denotes the proper vector composition of $\mathit{SE}(3) \times \mathbb{R}^n$.
    \item $\nabla(.)$ is the derivative operator with respect to (w.r.t.) corresponding variable, and $I$ is the identity matrix with the proper dimension.
    \item Throughout the paper, we use small letters to specify scalars and scalar-valued functions, bold small letters for vectors and vector-valued functions, and capital letters for matrices.
\end{itemize}
\subsection{Floating-Base Dynamics}
The dynamics of a floating-base system can be written as 
\begin{equation}\label{eq:rigid_body}
    M(\gencoor) \dot \genvel + \nonlinear(\gencoor,\genvel) = B \torque + J_c^\top \force,
\end{equation}
where $M \in \mathbb{R}^{(n+6) \times (n+6)}$ is the mass-inertia matrix, $\gencoor \in \mathit{SE}(3) \times \mathbb{R}^n$ denotes the configuration space, $\genvel \in \mathbb{R}^{n+6}$ encodes the vector of generalized velocities (or more precisely quasi-velocities \cite{baruh1999analytical} \S7.6), and $\nonlinear \in \mathbb{R}^{n+6}$ is a concatenation of nonlinear terms including centrifugal, Coriolis and gravitational effects. $B \in \mathbb{R}^{(n+6) \times n}$ is a selection matrix that separates the actuated and unactuated Degrees of Freedom (DoFs), $\torque \in \mathbb{R}^n$ is the vector of actuated joint torques, $J_c \in \mathbb{R}^{3m \times (n+6)}$ is the Jacobian of $m$ feet in contact, and finally $\force \in \mathbb{R}^{3m}$ is the vector of contact forces (here we assume point-contact feet for quadrupeds, however all results hold for humanoids with flat feet as well).
\subsection{Motion and Constraint Dynamics}
Assuming rigid non-slipping contact with the environment, and using the orthogonal projection operator $N = I - J_c^{\dagger} J_c$ (where $^{\dagger}$ is the Moore-Penrose inverse) which projects into the nullspace of the contact Jacobian, we can divide \eqref{eq:rigid_body} into the following set of equations \cite{mistry2012operational}
\begin{subequations}
\label{eq:rigid_body_separation}
\begin{align}
    NM(\gencoor) \dot \genvel + N\nonlinear(\gencoor,\genvel) &= NB \torque, \label{subeq:motion_space}\\
    (I-N)(M(\gencoor) \dot \genvel + \nonlinear(\gencoor,\genvel)) &= (I-N)B \torque + J_c^\top \force. \label{subeq:constraint_space}
\end{align}
\end{subequations}
Equation \eqref{subeq:motion_space} encodes the dynamics of the robot motion independent of the constraint forces, while \eqref{subeq:constraint_space} yields the dynamics of the system in the contact constraint space \cite{mistry2012operational}. In other words, $NB \torque$ is used to move the system on a desired trajectory without violating the contact constraint, while $(I-N)B \torque$ preserves the stationary contact. Interestingly, since \eqref{subeq:motion_space} is independent of contact forces, we can use it to relate the measured joint torques to the joint accelerations without knowledge of the contact forces. This is especially useful for agile legged robots whose endeffectors are often not equipped with contact force/torque sensors.

Since the feet in contact are assumed stationary, we can write down the following constraint and differentiate both sides w.r.t. time, yielding
\begin{equation}\label{eq:contact_constraint}
    (I-N)\genvel=\boldsymbol{0} \Rightarrow (I-N) \dot \genvel = \dot N \genvel.
\end{equation}
Defining the constraint-consistent mass-inertia matrix as $M_c = NM + I - N$ and substituting it along with \eqref{eq:contact_constraint} into \eqref{subeq:motion_space}, we have
\begin{align}
    M_c(\gencoor) \dot \genvel - \dot N \genvel + N\nonlinear(\gencoor,\genvel) = NB \torque. \label{eq:motion_space_invertible}
\end{align}
Now, we can invert $M_c$ and solve the constraint-consistent equations of motion for $\dot \genvel$ to yield
\begin{align}
    \dot \genvel = M_c^{-1} (\dot N \genvel - N\nonlinear + NB \torque). \label{eq:motion_space_acceleration}
\end{align}
\subsection{Relation between Joint Torques and Centroidal States}
Denoting the centroidal momentum vector by $\centroidal \in \mathbb{R}^6$, we can write down its relation with generalized velocities using
\begin{align}\label{eq:centroidal_momentum}
    \centroidal = \begin{bmatrix}
    \lmom,\amom
    \end{bmatrix}^\top = A_G(\gencoor) \genvel,
\end{align}
where $A_G(\gencoor)$ is the Centroidal Momentum Matrix \cite{orin2008centroidal}. Taking the derivative of \eqref{eq:centroidal_momentum} w.r.t. time yields
\begin{align}\label{eq:centroidal_momentum_derivative}
    \centroidalrate = A_G(\gencoor) \dot\genvel + \dot A_G(\gencoor) \genvel.
\end{align}
Substituting \eqref{eq:motion_space_acceleration} into \eqref{eq:centroidal_momentum_derivative}, we have
\begin{align}\label{eq:process_model}
    \centroidalrate = A_G(\gencoor) (M_c^{-1} (\dot N \genvel - N\nonlinear + NB \torque)) + \dot A_G(\gencoor) \genvel.
\end{align}
Assuming that $\gencoor, \, \genvel$ can be measured using joint encoders, and the base state is estimated separately using a base estimator, we can write down the momentum dynamics in the following compact form:
\begin{align}\label{eq:process_model_simplified}
    \centroidalrate = D(\gencoor, \, \genvel) \torque + \boldsymbol{b}(\gencoor, \, \genvel),
\end{align}
where
\begin{subequations}
    \begin{align}\label{eq:process_model_simplified_components}
        D(\gencoor, \, \genvel) &\overset{\Delta}{=} A_G(\gencoor) M_c^{-1} N B,\\
        \boldsymbol{b}(\gencoor, \, \genvel) &\overset{\Delta}{=} A_G(\gencoor) M_c^{-1} (\dot N \genvel - N\nonlinear) + \dot A_G(\gencoor) \genvel.
    \end{align}
\end{subequations}
\section{Centroidal State Estimation}
We aim to estimate the centroidal states $\boldsymbol{x} \in \mathbb{R}^9 = \begin{bmatrix}
\com, \lmom, \amom
\end{bmatrix}^\top$ using joint torque measurements and the above centroidal momentum dynamics. The process model is
\begin{align}\label{eq:estiamtion_process_model}
     \comrate &= \frac{1}{m} \lmom, \nonumber\\
    \begin{bmatrix}
{\lmomrate}, {\amomrate}
\end{bmatrix}^\top &= D \hat{\torque} + \boldsymbol{b},
\end{align}
where $\hat{\torque}$ is the measured joint torques, $m$ is the total mass of the robot, $\com \in \mathbb{R}^3$ is the center of mass (CoM) position, and $\lmom \in \mathbb{R}^3$ and $\amom \in \mathbb{R}^3$ are the linear and angular components of centroidal momentum respectively. These values are computed using the results of base state estimation, measured generalized joint states, kinematics, and inertial properties of the robot. Specifically, we use \eqref{eq:centroidal_momentum} to compute the momenta while the CoM position is computed using the estimated base state and joint measurements as
\begin{equation}\label{eq:com}
    \com = \boldsymbol{g}(\gencoor).
\end{equation}
The measurement model is
\begin{align} \label{eq:measurement_model}
    &\boldsymbol{y} : (\mathit{SE}(3) \times \mathbb{R}^n, \; \mathbb{R}^{n+6}) \mapsto \mathbb{R}^9, \nonumber \\
    &\boldsymbol{y}(\gencoor, \genvel) = 
    \begin{bmatrix}
        \boldsymbol{g}(\gencoor), \centroidal
    \end{bmatrix}^\top.
\end{align}
As shown in \cite{rotella2015humanoid}, although the centroidal states can be computed directly, they are subject to considerable noise and modeling errors. By fusing the computed states with a torque-based process model, we similarly obtain low-noise estimates with minimal delay which is suitable for high-bandwidth control. 
\begin{remark}
    It is important to note that the right hand side of the process models \eqref{eq:estiamtion_process_model} is an implicit function of $\com\,, \lmom$, and $\amom$. Hence, we need to compute its Jacobian to be used inside the EKF formulation. We show in Sec. \ref{sec:linearization} how this Jacobian can be computed. 
    One simpler approach than computing these Jacobians and using EKF would be to treat \eqref{eq:estiamtion_process_model} as a time-varying linear model where $D$ and $\boldsymbol{b}$ are evaluated at each configuration. This would be a special case of the more general EKF framework we are proposing in this paper.
\end{remark}
\begin{remark}
For both base and centroidal state estimations, a contact detection mechanism for each foot is employed. One might ask since we estimate the contact forces for contact detection, then why not use a centroidal state estimator based on estimated contact forces (for instance as in \cite{rotella2015humanoid}) which would render the process model much simpler compared to \eqref{eq:estiamtion_process_model}? The answer to this question is threefold: 1) contact detection is normally performed by setting a threshold only on the \textit{estimated normal force} separately for each leg neglecting the dynamics. This force measurement is not accurate enough to be used in the centroidal state estimation problem. 2) Instead of using the whole-body dynamics to estimate the contact forces and then use it to estimate the centroidal states, our approach estimates those quantities directly from torque measurements. 3) Many legged robots are endowed with binary contact sensors at their endeffectors \cite{grimminger2020open,chignoli2021humanoid} that can directly detect contact events, hence there is no need to estimate endeffector forces.
\end{remark}
\subsection{Extended Kalman Filtering}
We choose to implement our estimator as an Extended Kalman Filter (EKF). The EKF estimates the mean $\boldsymbol{\mu}$ and the error covariance matrix ${P}$ over the state $\boldsymbol{x}_k$, where both process and measurement models are corrupted by additive zero-mean Gaussian noise. The discrete-time stochastic system evolves as:
\begin{subequations}
\begin{align}
    \boldsymbol{x}_k &= \boldsymbol{f}(\boldsymbol{x}_{k-1},
                        \boldsymbol{u}_{k}) + \boldsymbol{w}_k,\\
    \boldsymbol{z}_k &= \boldsymbol{h}(\boldsymbol x_k) + \boldsymbol{v}_k,
\end{align}
\end{subequations}
where the input $\boldsymbol{u}_k = \torque_k$ is the vector of joint torques and $\boldsymbol{z}_k$ is the measurement. The process noise $\boldsymbol{w}_k \sim \mathcal{N}(\boldsymbol{0}, {Q}_k)$ and measurement noise $\boldsymbol{v}_k \sim \mathcal{N}(\boldsymbol{0}, {R}_k)$ are parameterized by the corresponding noise covariance matrices ${Q}_k$ and ${R}_k$. These matrices constitute the main tuning parameters of the filter. Note that the process noise is chosen to be purely additive (rather than based on the uncertainty in torque sensing) for the sake of simplicity in filter implementation. In other words, we represent the effects of joint torque measurements and dynamic model errors as additive noise in the process and measurement models respectively.

In the prediction step, having determined the set of feet in contact using estimated contact forces, the mean is propagated using the discrete-time nonlinear process model, and the process model Jacobian ${F}_k =  \left.\frac{{\partial{\boldsymbol{f}(\boldsymbol{x})}}}{\partial{\boldsymbol{x}}}\right\vert_{\boldsymbol{x}=\boldsymbol{\mu}_{k}^{+}}^{}$ is used to propagate the error covariance matrix as
\begin{subequations}
\begin{align}\label{eq:prediction_step}
    \boldsymbol{\mu}_k^- &= \boldsymbol{f}(\boldsymbol{\mu}_{k-1}^+,\boldsymbol{u}_{k}),\\
    {P}_k^- &= ({F}_{k-1} {P}_{k-1}^+ {F}_{k-1}^T) 
                        + {Q}_k,\label{eq:covariance propagation}
\end{align}
\end{subequations}
where the minus superscript denotes the \emph{a priori} (before measurement update) estimate, and the plus superscript denotes the \emph{a posteriori} (after measurement update).

In the update step, the measurements $\boldsymbol{z}_k$ are integrated into the EKF using the \emph{a priori} estimates and the measurement noise covariance as follows: 
\begin{subequations}
\begin{align}\label{eq:update_step}
    {K}_k &= {P}_k^- {H}_k^T 
                        ({H}_k {P}_k^- {H}_k^T + {R}_k)^{-1},\\
    \boldsymbol{\mu}_k^+ &= \boldsymbol{\mu}_k^- + {K}_k  (\boldsymbol{z}_k -
                            \boldsymbol{h}(\boldsymbol{\mu}_k^-)),\\
    {P}_k^+ &= ({I} - {K}_k {H}_k) {P}_{k}^{-},
\end{align}
where $K_k$ is the Kalman gain, and $H_k = \left.\frac{{\partial{\boldsymbol{h}(\boldsymbol{x})}}}{\partial{\boldsymbol{x}}}\right\vert_{\boldsymbol{x}=\boldsymbol{\mu}_{k}^{-}}$ is the measurement Jacobian matrix.
\end{subequations}
\subsection{Discrete-Time Nonlinear Process Model}
In the prediction step, the mean of the state is propagated using the discretized nonlinear process model. We discretize (\ref{eq:estiamtion_process_model}) using a first-order explicit Euler integration as follows:
\begin{align}
    \boldsymbol{\mu}_{k} &=
    \begin{bmatrix}
        \boldsymbol{c}_{G[k-1]}\\
        \boldsymbol{l}_{G[k-1]}\\
        \boldsymbol{k}_{G[k-1]}\\
    \end{bmatrix} +
    \begin{bmatrix}
        \frac{1}{m} \boldsymbol{l}_{G[k-1]} \Delta t\\
        \\
        \begin{bmatrix}
                D \hat{\torque}_{k} + \boldsymbol{b}\\
        \end{bmatrix} \Delta t
    \end{bmatrix}, 
\end{align}
where $\Delta t$ is the discretization time step, and $\hat{\torque}_{k}$ are the measured joint torques.
\subsection{Discrete-Time Linearized Process Model}\label{sec:linearization}
In order to propagate the state covariance in the prediction step \eqref{eq:covariance propagation}, we first linearize then discretize the dynamics at the filtering frequency. Since the nonlinear dynamic model of the momentum states $[\lmom, \amom]^\top$ is expressed implicitly as a function of the centroidal states, we resort to finite differencing for the computation of the continuous-time process model Jacobian ${F}_c$ in (\ref{eq:process_model_simplified}).

To compute the partial derivative of $\centroidalrate$ w.r.t. $c_x$, we first find the deviation of $\delta \gencoor$ corresponding to the variation of CoM position in the $x$-direction. By considering the first three rows (linear momentum part) of $A_G(\gencoor)$ as $A_G^{\prime}(\gencoor)$, we can write down the variation of the CoM position as a function of variation of generalized coordinates as
\begin{align}
    \delta \com = \frac{1}{m} A_G^{\prime}(\gencoor) \delta \gencoor.
\end{align}
Finding the variation of $\gencoor$ only in the $x$-direction of the CoM position simply becomes
\begin{align}
    \delta \gencoor = m A_G^{' \dagger}(\gencoor)\begin{bmatrix} \delta c_x,\, 0,\, 0
    \end{bmatrix}^\top.
\end{align}
Using $\gencoor^+ = \gencoor \oplus \delta \gencoor$, we can find the part of the Jacobian matrix corresponding to $c_x$ as
\begin{figure*}[htb]
	\centering
	\includegraphics[height=0.10\linewidth, trim={0cm 0cm 0cm 0cm}, clip]{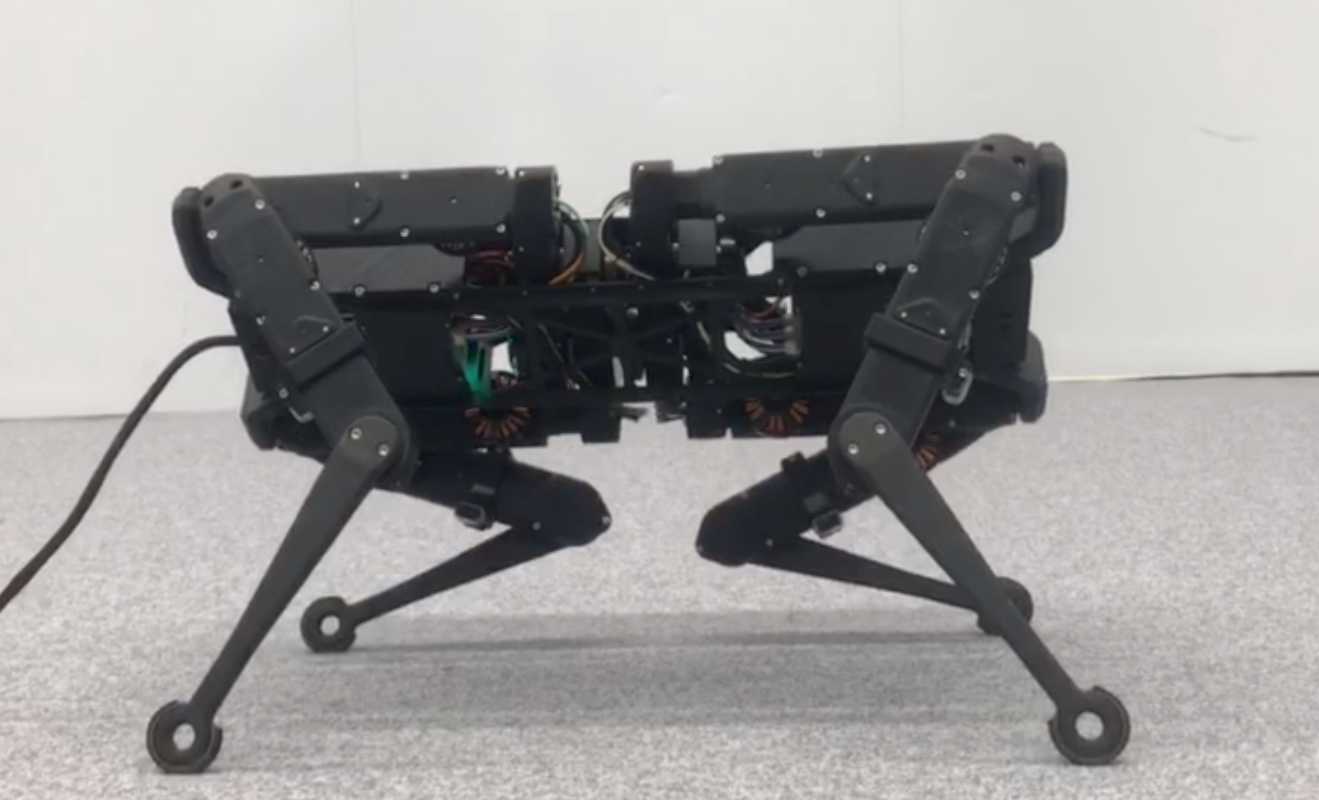}%
	\includegraphics[height=0.10\linewidth, trim={0cm 0cm 0cm 0cm}, clip]{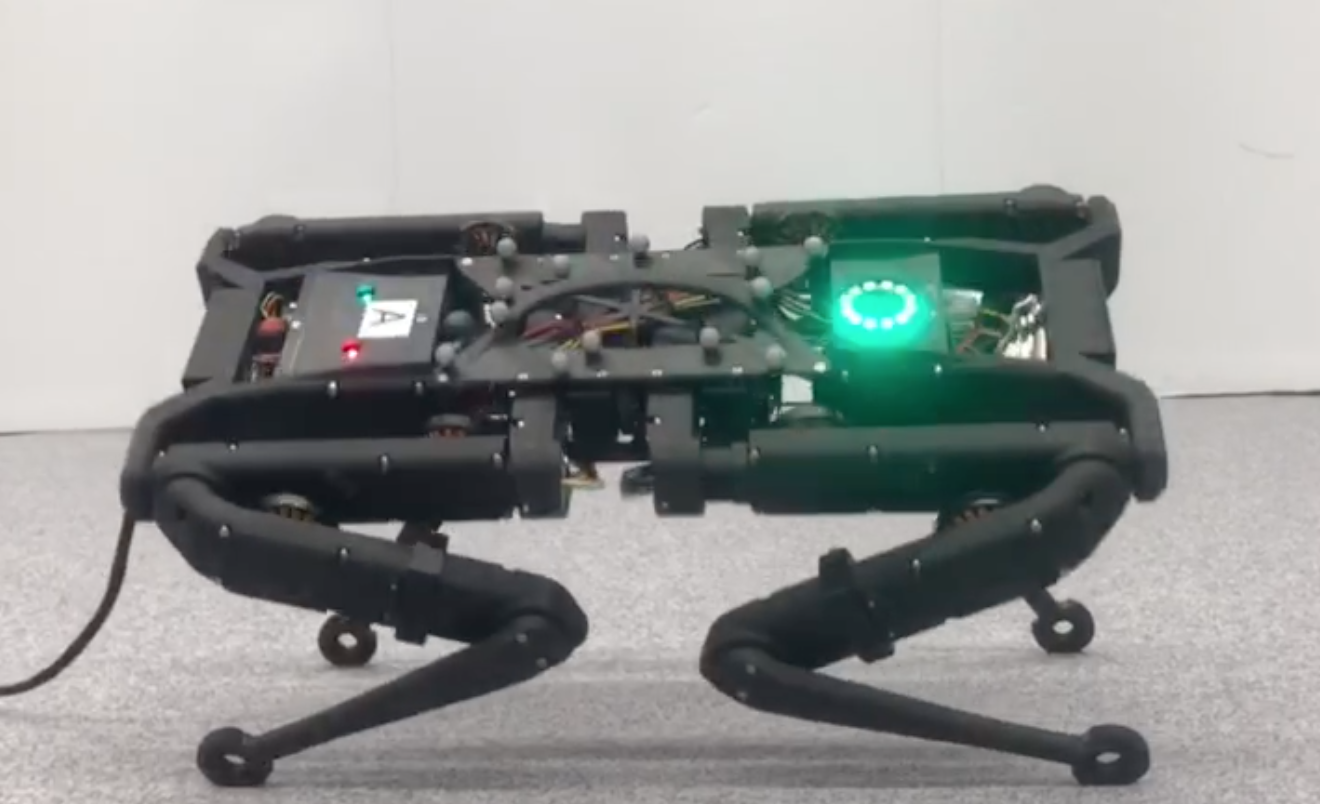}%
	\includegraphics[height=0.10\linewidth, trim={0cm 0cm 0cm 0cm}, clip]{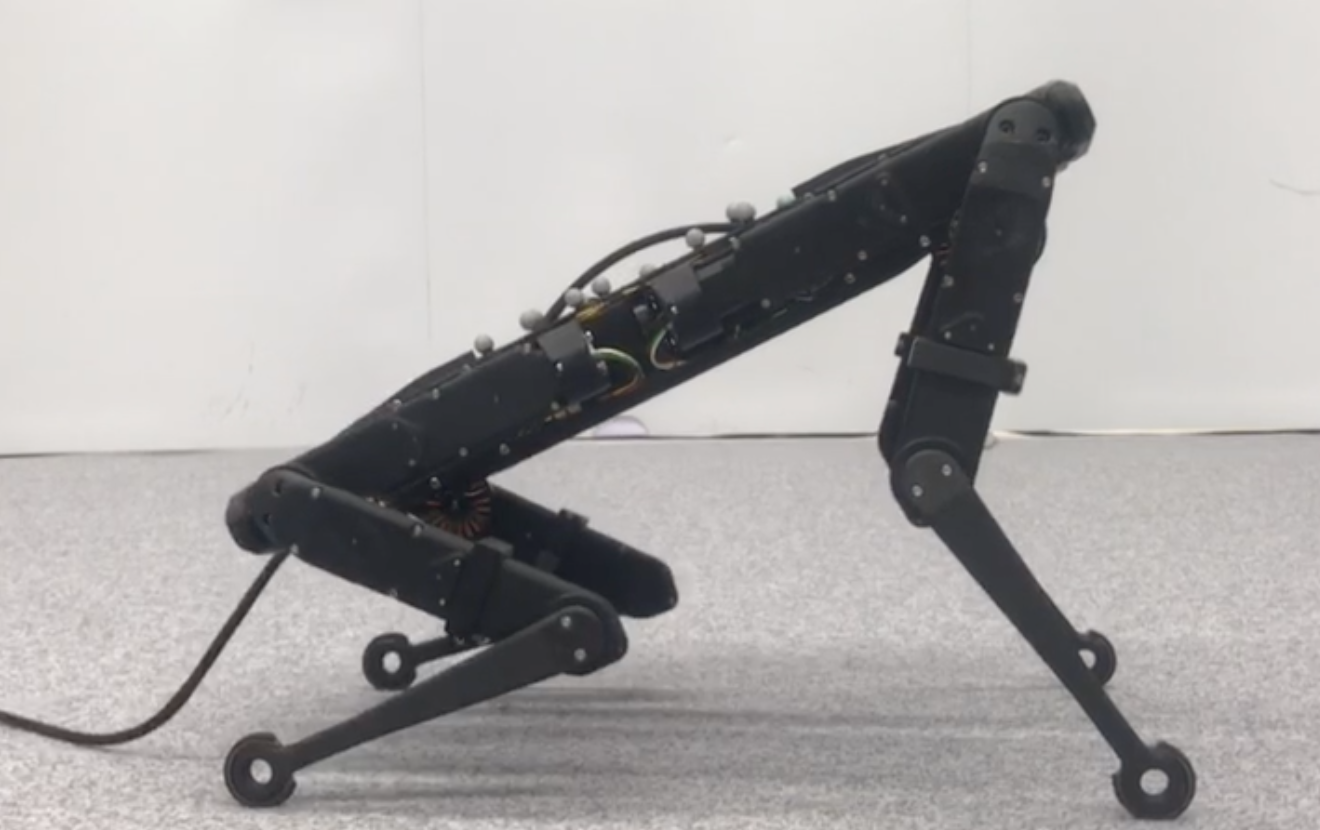}%
	\includegraphics[height=0.10\linewidth, trim={0cm 0cm 0cm 0cm}, clip]{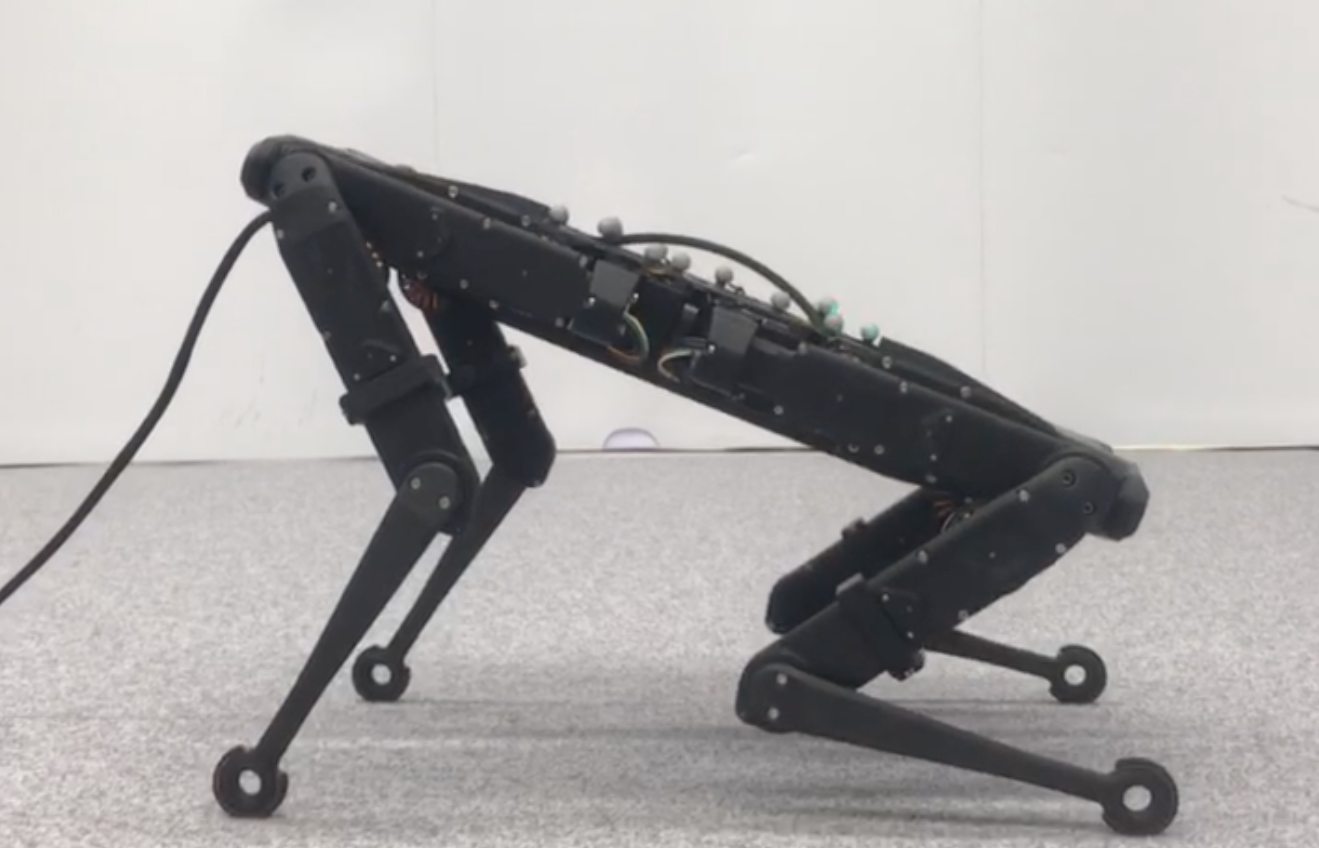}%
	\includegraphics[height=0.10\linewidth, trim={0cm 0cm 0cm 0cm}, clip]{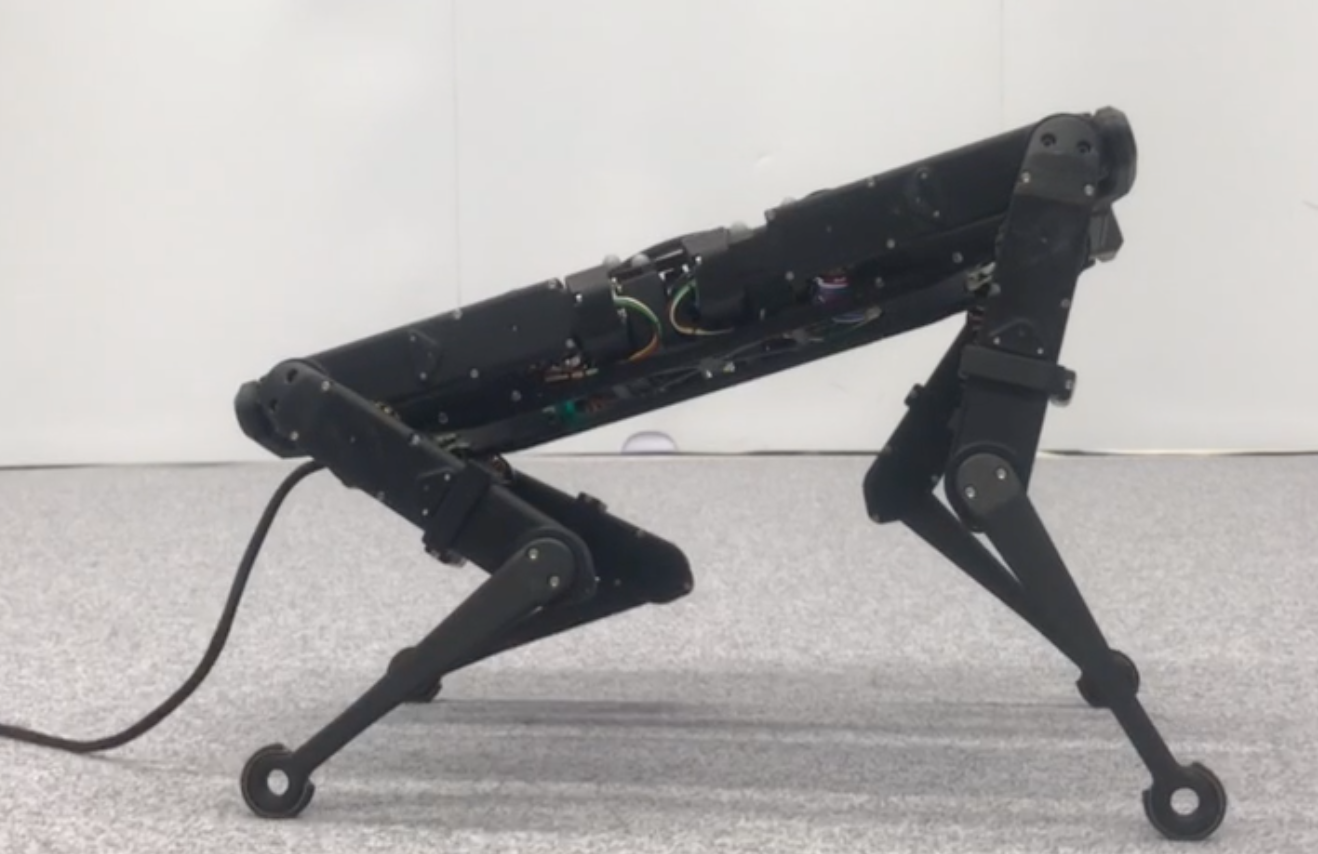}%
	\includegraphics[height=0.10\linewidth, trim={0cm 0cm 0cm 0cm}, clip]{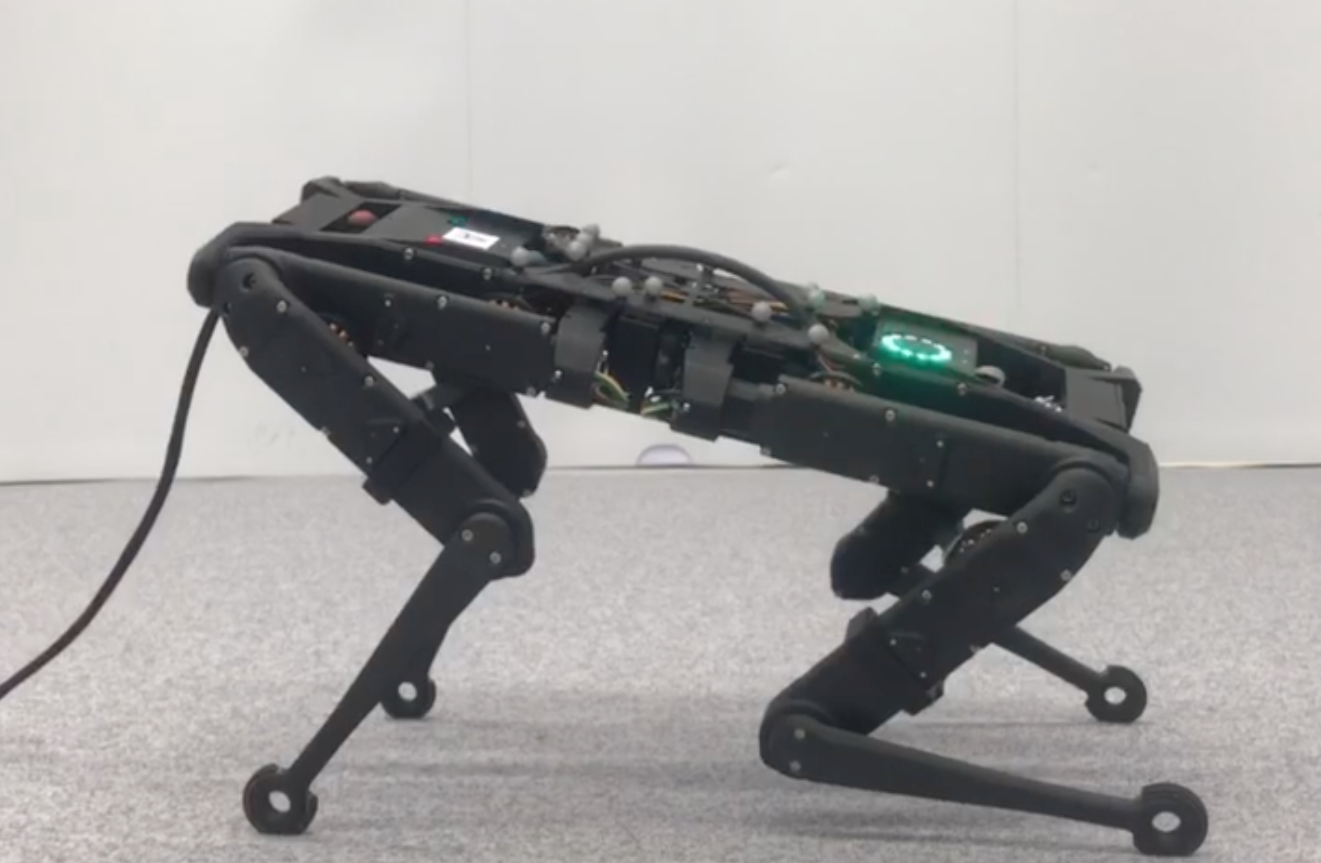}%
	\caption[]{Snapshots of the first motion scenario (moving the base while balancing, with all feet in contact).}
	\label{fig:base_snapshots}
	\vspace{-4.5mm}
\bigskip
    \scalebox{0.85}{\input{files/rotation.pgf}}
    \caption{Motion \#1: Moving the base while balancing. Estimation of the CoM position $(m)$ (top row), linear momentum $(\frac{kgm}{s})$ (middle row) and angular momentum $(\frac{kgm^2}{s})$ (bottom row); computed centroidal states in \textcolor{blue}{blue}, and EKF in \textcolor{red}{red}.}
    \label{fig:moving_base}
    \vspace{-5mm}
\end{figure*}
%
\begin{subequations}
\begin{align}
    \Delta \centroidalrate &= \centroidalrate(\gencoor^+, \genvel, \hat{\torque}) - \centroidalrate(\gencoor, \genvel, \hat{\torque}),\\
    \Delta \com &= \boldsymbol{g}(\gencoor^+) - \boldsymbol{g}(\gencoor), \label{eq:Delta com}\\
      \frac{\partial \centroidalrate}{\partial c_x} &= \frac{\Delta \centroidalrate}{\Delta c_x},
\end{align}
\end{subequations}
where $\Delta c_x \in \mathbb{R}$ is the $x$ component of \eqref{eq:Delta com}. Following the same procedure above, we compute the derivative w.r.t. $c_y$ and $c_z$ by considering small variation as $[0, \delta c_y, 0]^\top$ and $[0, 0, \delta c_z]^\top$ respectively.\\
Similarly, to find the derivative of $\centroidalrate$ w.r.t. linear momentum $l_x$, we compute the deviation of $\delta \genvel$ corresponding to the variation of linear momentum only in the $x$-direction. To do so, we rewrite \eqref{eq:centroidal_momentum_derivative} as
\begin{align}
    \delta \centroidal = A_G(\gencoor) \delta \genvel + \delta A_G(\gencoor) \genvel.
\end{align}
By considering $A_G(\gencoor)$ constant for small variations (i.e. $\delta A_G(\gencoor) = 0$), we can find a small variation of $\genvel$ corresponding to the small variation of linear momentum in the x-direction  being
\begin{align}
 \delta \genvel = A_G^{\dagger}(\gencoor)  \begin{bmatrix}
 \delta l_x, 0, 0, 0, 0, 0\end{bmatrix}^\top.
\end{align}
Given $\genvel^+ = \genvel + \delta \genvel$, we compute the part of Jacobian matrix corresponding to  $l_x$ as
\begin{subequations}
\begin{align}
    \Delta \centroidalrate &= \centroidalrate(\gencoor, \genvel^+, \hat{\torque}) - \centroidalrate(\gencoor, \genvel, \hat{\torque}),\\
    \Delta \lmom &= \lmom(\gencoor, \genvel^+) - \lmom(\gencoor, \genvel), \label{eq:Delta lin. mom}\\
     \frac{\partial \centroidalrate}{\partial l_x} &= \frac{\Delta \centroidalrate}{\Delta l_x},
    \end{align}
\end{subequations}
where $\Delta l_x \in \mathbb{R}$ is the $x$ component of \eqref{eq:Delta lin. mom}. Following the same procedure above, we compute the rest of the remaining partial derivatives of the linear and angular momenta w.r.t. $l_y, l_z, k_x, k_y$, and $k_z$. Finally, we assemble the continuous-time process model Jacobian matrix $F_c \in \mathbb{R}^{9\times9}$ as
%
\begin{align}
    F_c = 
    \begin{bmatrix}
    0_{3 \times 3} \qquad \frac{1}{m} I_{3 \times 3} \qquad 0_{3 \times 3}\\
    \Omega_{6 \times 3} \qquad
    \Lambda_{6 \times 3} \qquad
    \Gamma_{6 \times 3}
  \end{bmatrix},                
\end{align}
where $\Omega \triangleq \begin{bmatrix}
        \frac{\Delta \centroidalrate}{\Delta \boldsymbol{c}_x}, \frac{\Delta \centroidalrate}{\Delta \boldsymbol{c}_y}, \frac{\Delta \centroidalrate}{\Delta \boldsymbol{c}_z}
    \end{bmatrix}$, $\Lambda \triangleq \begin{bmatrix}
        \frac{\Delta \centroidalrate}{\Delta \boldsymbol{l}_x}, \frac{\Delta \centroidalrate}{\Delta \boldsymbol{l}_y}, \frac{\Delta \centroidalrate}{\Delta \boldsymbol{l}_z} 
    \end{bmatrix}$, $\Gamma \triangleq \begin{bmatrix}
        \frac{\Delta \centroidalrate}{\Delta \boldsymbol{k}_x}, \frac{\Delta \centroidalrate}{\Delta \boldsymbol{k}_y}, \frac{\Delta \centroidalrate}{\Delta \boldsymbol{k}_z}\\
    \end{bmatrix}$.
In the discrete-time form, the Jacobian is truncated at its first-order Taylor approximation for simplicity, yielding ${F}_k \approx  I_{9 \times 9} + {F}_c \Delta t$, and the discretized process noise covariance matrix is likewise approximated as ${Q}_k \approx {F}_k {Q}_c {F}_k^T \Delta t$.
Similarly, we can find the measurement Jacobian in the update step, by taking corresponding derivatives w.r.t. our state. By following the procedure described for computing prediction Jacobian, and defining proper variations for $\gencoor$, and $\genvel$ we get ${H}_k = I_{9 \times 9}$, which is consistent with the result of \cite{rotella2015humanoid}. We highlight that the measurement Jacobian is the identity matrix since the state is  directly measured.
\begin{figure*}[htb]
	\centering
	\includegraphics[height=0.1\linewidth, trim={0cm 0cm 0cm 0cm}, clip]{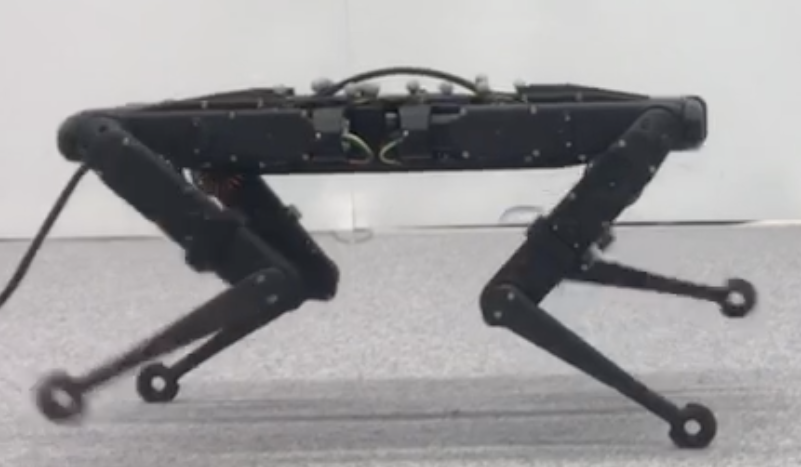}%
	\includegraphics[height=0.1\linewidth, trim={0cm 0cm 0cm 0cm}, clip]{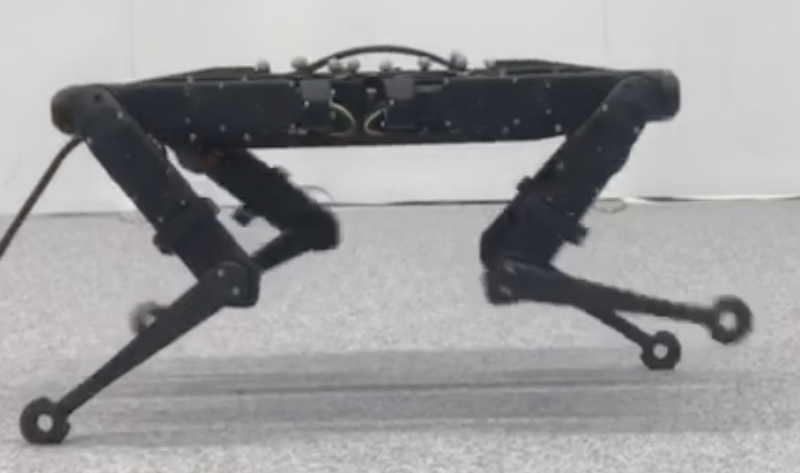}%
	\includegraphics[height=0.1\linewidth, trim={0cm 0cm 0cm 0cm}, clip]{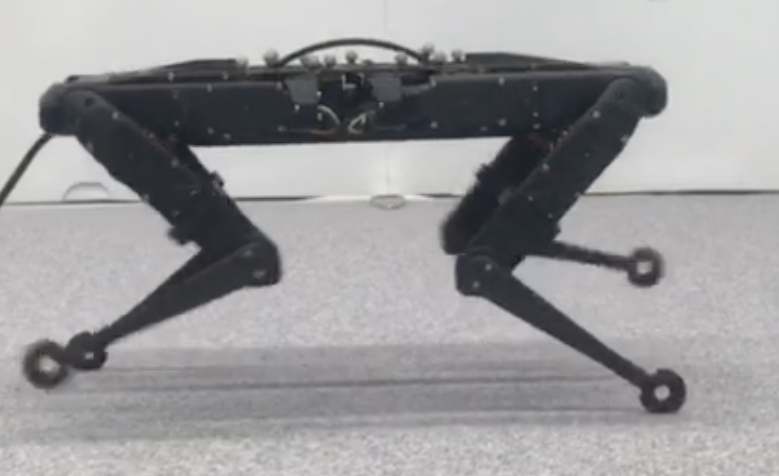}%
	\includegraphics[height=0.1\linewidth, trim={0cm 0cm 0cm 0cm}, clip]{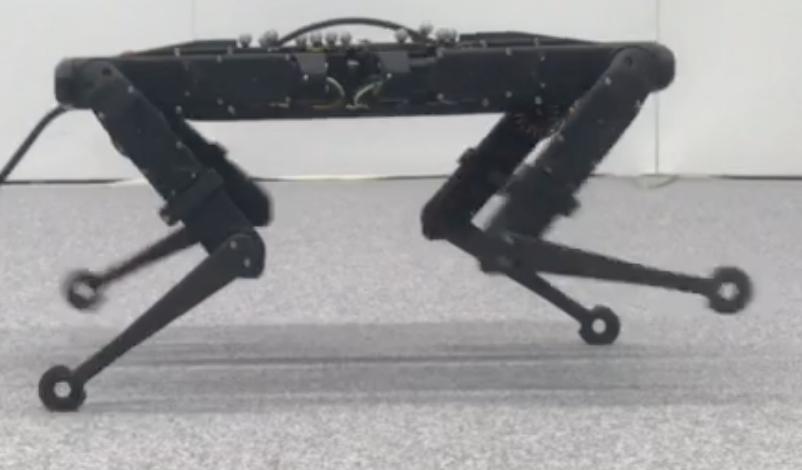}%
	\includegraphics[height=0.1\linewidth, trim={0cm 0cm 0cm 0cm}, clip]{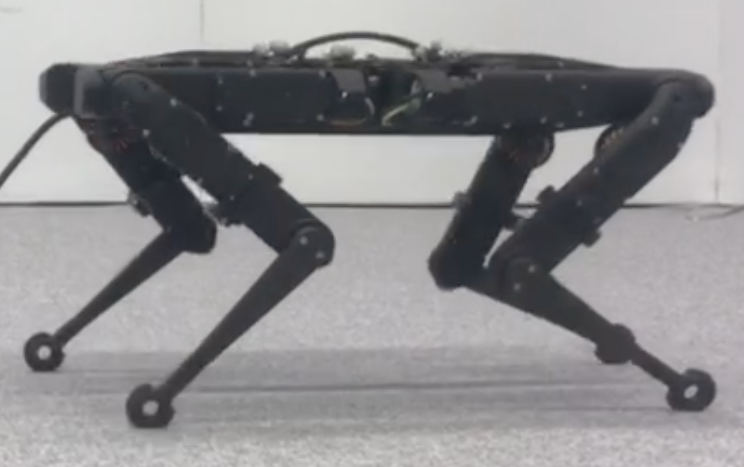}%
	\caption[]{Snapshots of the second motion scenario (forward trotting).}
	\label{fig:trot_snapshots}
	\vspace{-4.5mm}
\bigskip
    \scalebox{0.85}{\input{files/trot.pgf}}
    \caption{Motion \#2: Forward trotting with contact switching. Estimation of the CoM position $(m)$ (top row), linear momentum $(\frac{kgm}{s})$ (middle row) and angular momentum $(\frac{kgm^2}{s})$ (bottom row); computed centroidal states in \textcolor{blue}{blue}, and EKF in \textcolor{red}{red}.}
    \label{fig:trot}
    \vspace{-5mm}
\end{figure*}
This also validates the finite difference method proposed in this work for computing the process and measurement Jacobians.
\subsection{Observability Analysis}
The process model of the proposed estimator is a highly-nonlinear function of the states; in general, we may investigate the observability of a nonlinear estimator by forming the nonlinear observability matrix \cite{hermann1977observability}. 
The state of the estimator is observable if the observability matrix has full rank. In our case, the process model cannot be expressed explicitly in terms of the states, preventing us from computing Jacobians of $\boldsymbol{f}(\boldsymbol{x})$. However, because we directly measure the full state, we have $\nabla \boldsymbol{h}(\boldsymbol{x}) = {H}_k = I_{9 \times 9}$. This already means that the observability matrix has full column rank, so the state is observable. This is in agreement with the results of \cite{rotella2015humanoid}, where the observability matrix is computed for a similar momentum estimator. 
\section{Results}
In this section, we present the experimental results of performing estimation of the centroidal states both with and without the use of torque measurements, for a variety of different gaits on the quadruped robot Solo12 \cite{grimminger2020open}. Specifically, we consider three different behaviors: moving the base while balancing, trotting, and jumping. The trajectories are generated using the trajectory optimization framework in \cite{ponton2021efficient}, and the whole-body controller in \cite{grimminger2020open} is used to track these trajectories while satisfying friction cone constraints. The process and measurement models covariance matrices are tuned as $Q_k =$ $\Diag(10^{-7}, 10^{-7}, 10^{-7}, 10^{-5}, 10^{-5}, 10^{-5},\! 10^{-4},\! 10^{-4},\! 10^{-4})$ and $R_K = (10^{-5}) * I_9$ respectively. Note that we do not close the control loop around the estimated centroidal states; this is left to future work. Note also that the measured torques are computed by simply multiplying the measured current to the motor constant. While this does not give a precise torque measurement at the output of the joints due to the unmodeled drive system (e.g. friction, elasticity, etc.), we show in the following experiments that this is not an issue for the estimator. Note that since we cannot measure directly the centroidal states, we compute them using measurements from the motion capture system and joint encoders. Given these base and joint state measurements, we employ \eqref{eq:centroidal_momentum} and \eqref{eq:com} to construct the full centroidal states.
\subsection{Motion \#1: Moving the Base while Balancing}
In the first scenario we consider a wobbling motion of the base without contact switching (Fig. \ref{fig:base_snapshots}). The main goal of this test is to evaluate the quality of the estimated states in isolation from uncertainties introduced by contact switching as well as contact detection. 
Fig. \ref{fig:moving_base} shows the computed centroidal states in blue, compared against the estimated states from our proposed torque measurement-based EKF in red. As it can be clearly seen in this figure, the use of torque measurements in the proposed EKF significantly filter the centroidal momentum without adding delays.
\begin{figure*}[htb]
	\centering
	\includegraphics[height=0.12\linewidth, trim={0cm 0cm 0cm 0cm}, clip]{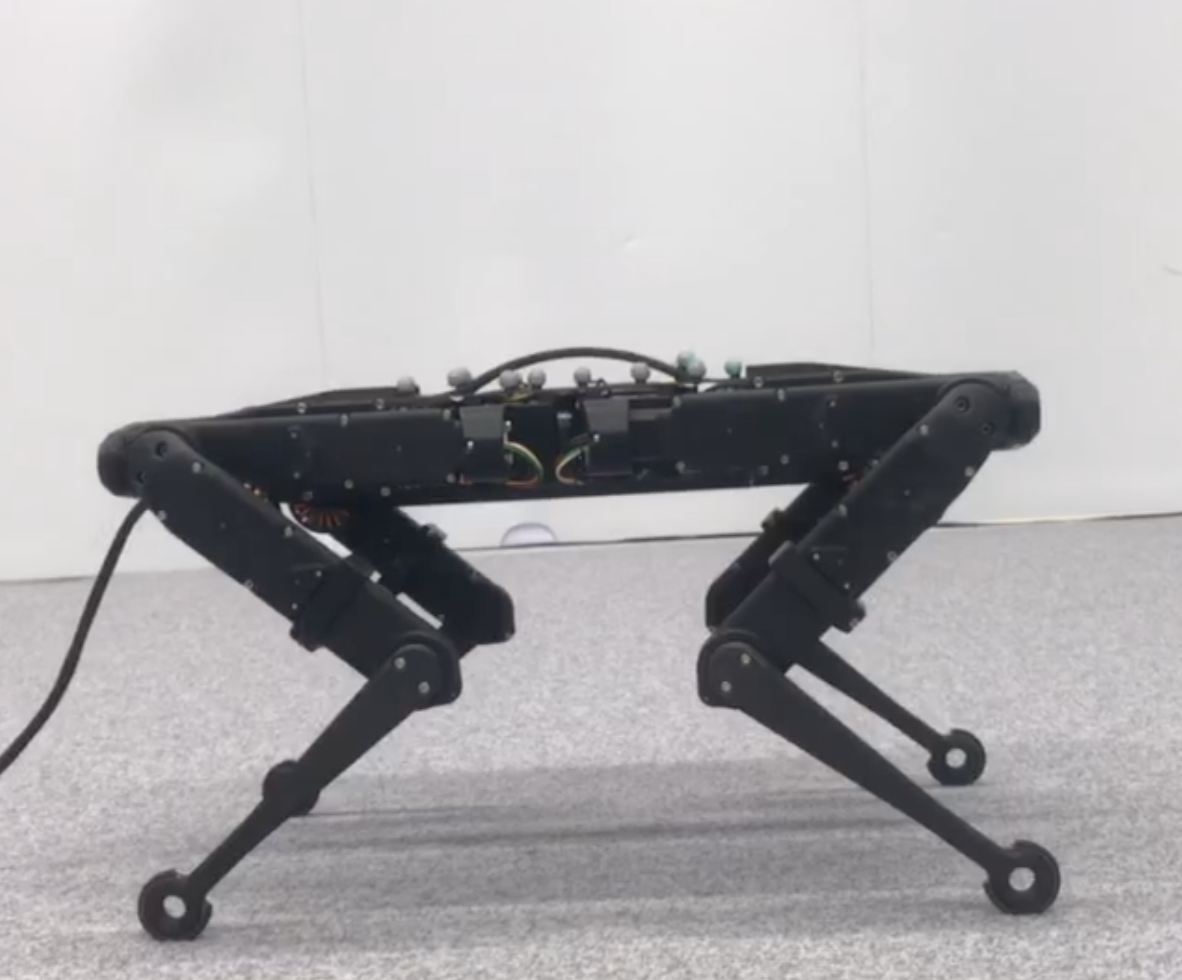}%
	\includegraphics[height=0.12\linewidth, trim={0cm 0cm 0cm 0cm}, clip]{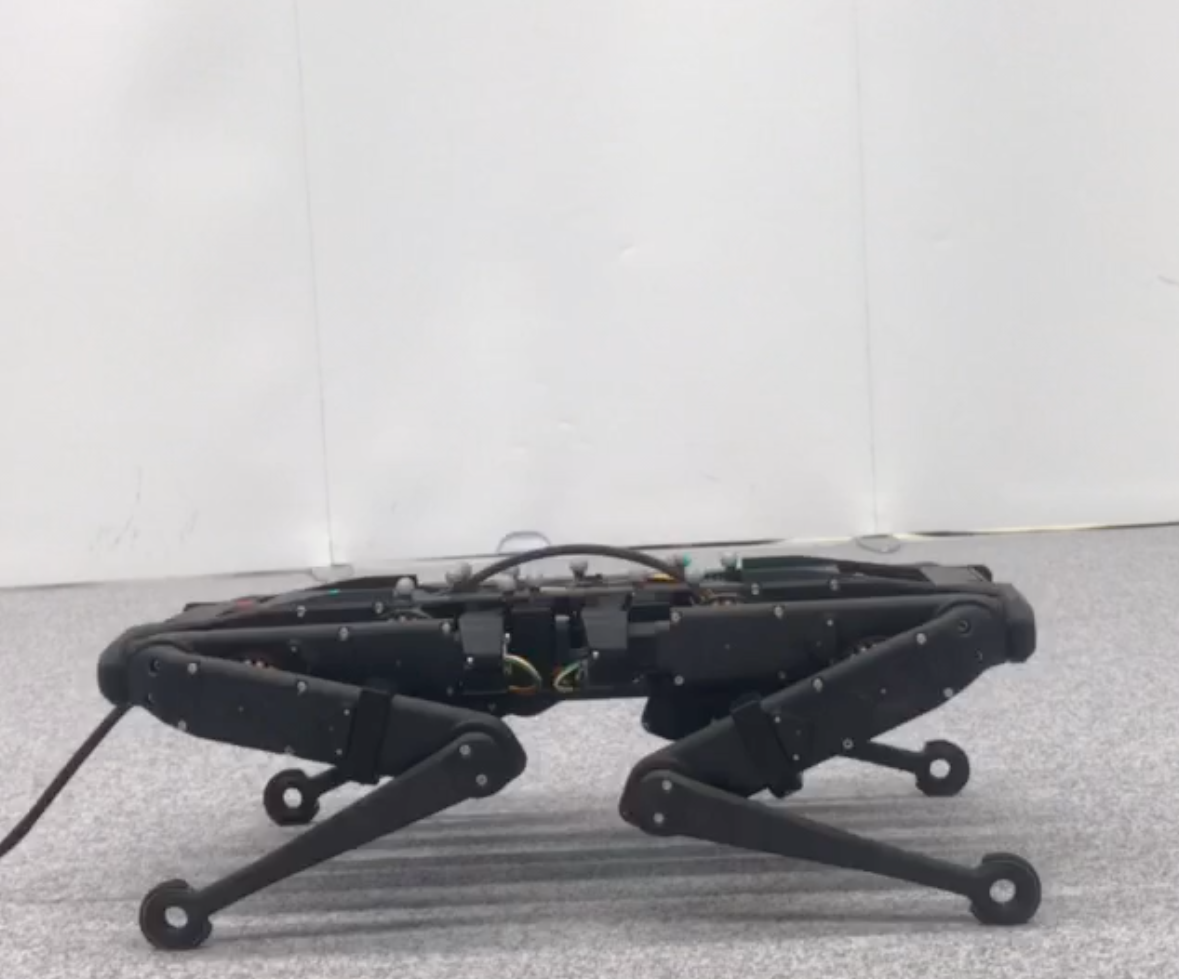}%
	\includegraphics[height=0.12\linewidth, trim={0cm 0cm 0cm 0cm}, clip]{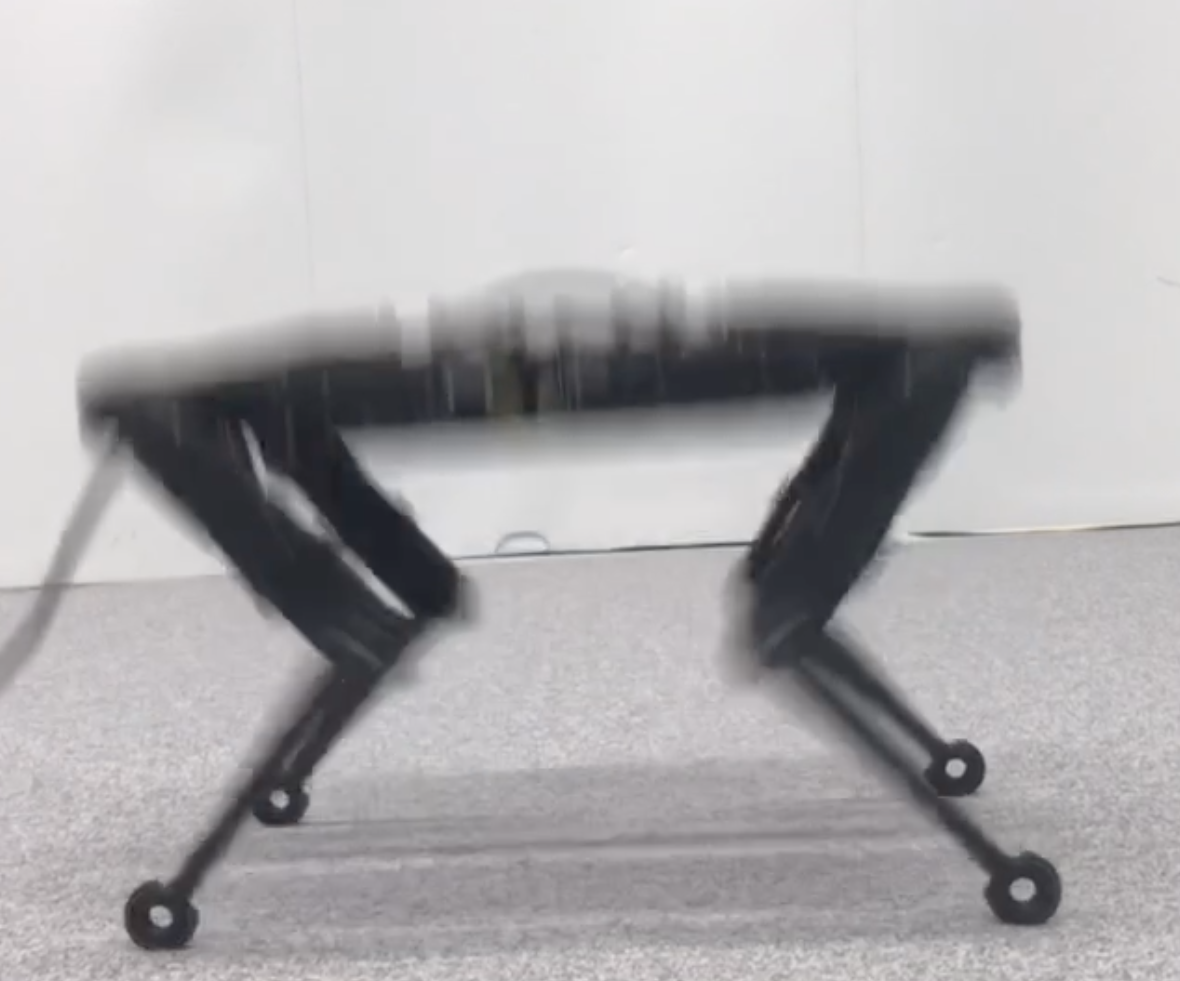}%
	\includegraphics[height=0.12\linewidth, trim={0cm 0cm 0cm 0cm}, clip]{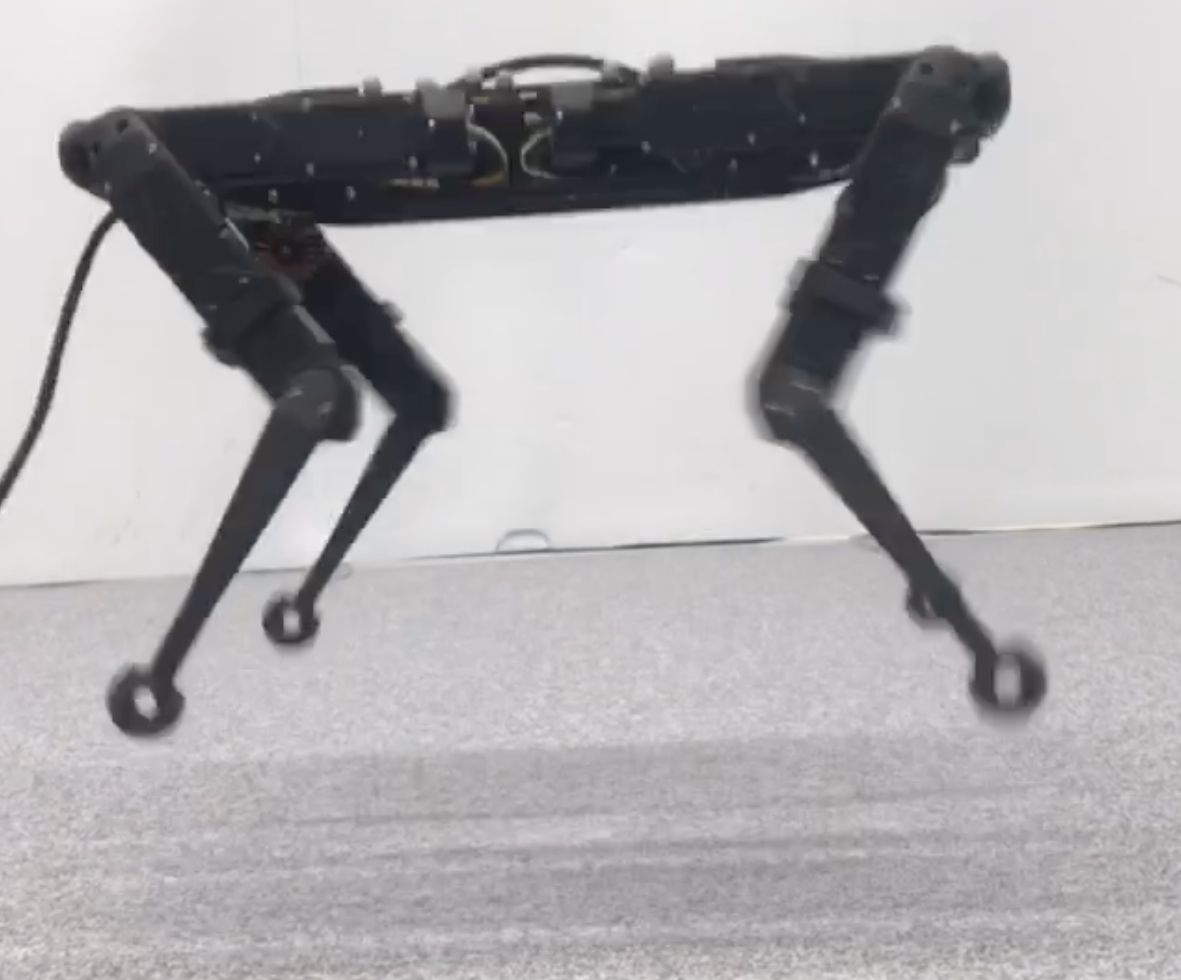}%
	\includegraphics[height=0.12\linewidth, trim={0cm 0cm 0cm 0cm}, clip]{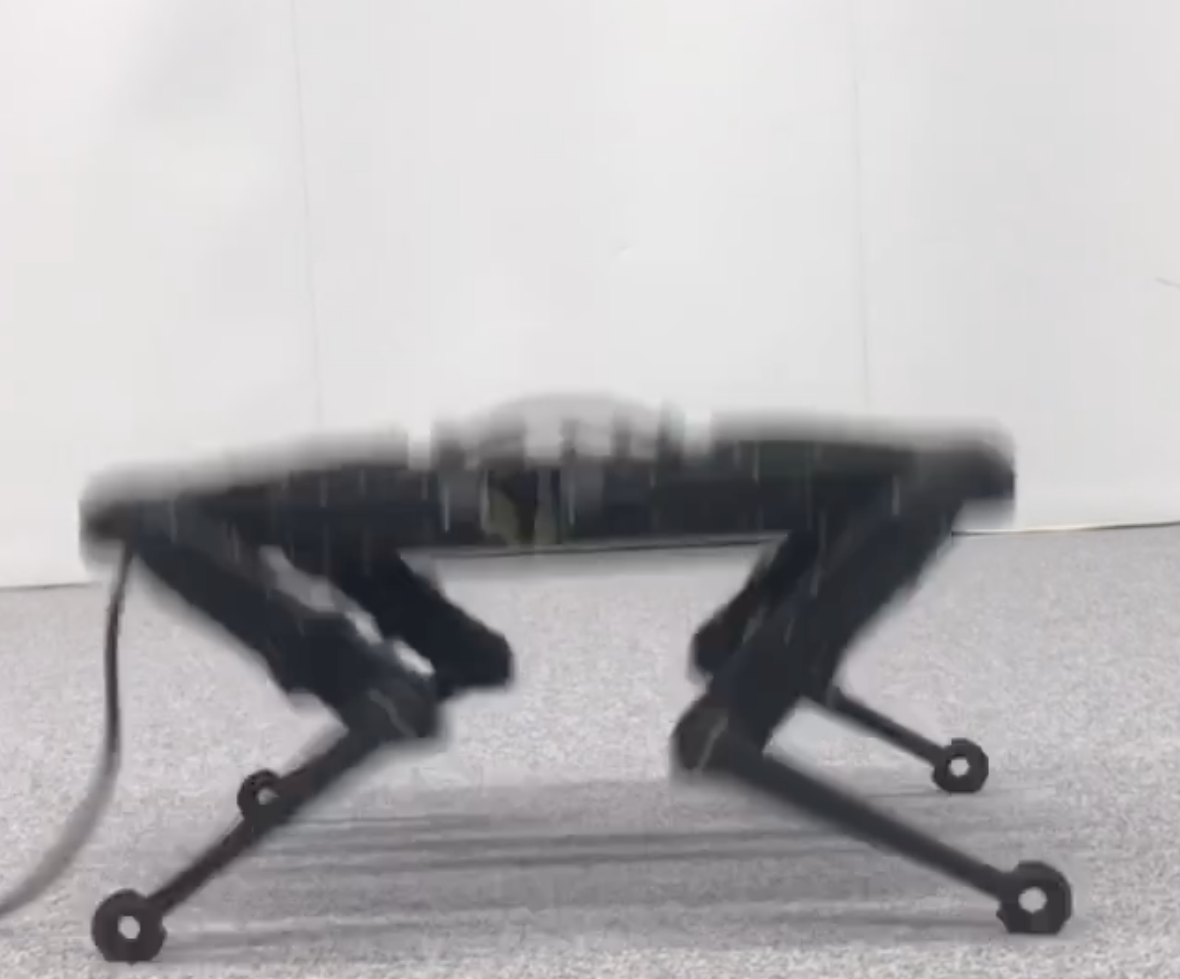}%
	\includegraphics[height=0.12\linewidth, trim={0cm 0cm 0cm 0cm}, clip]{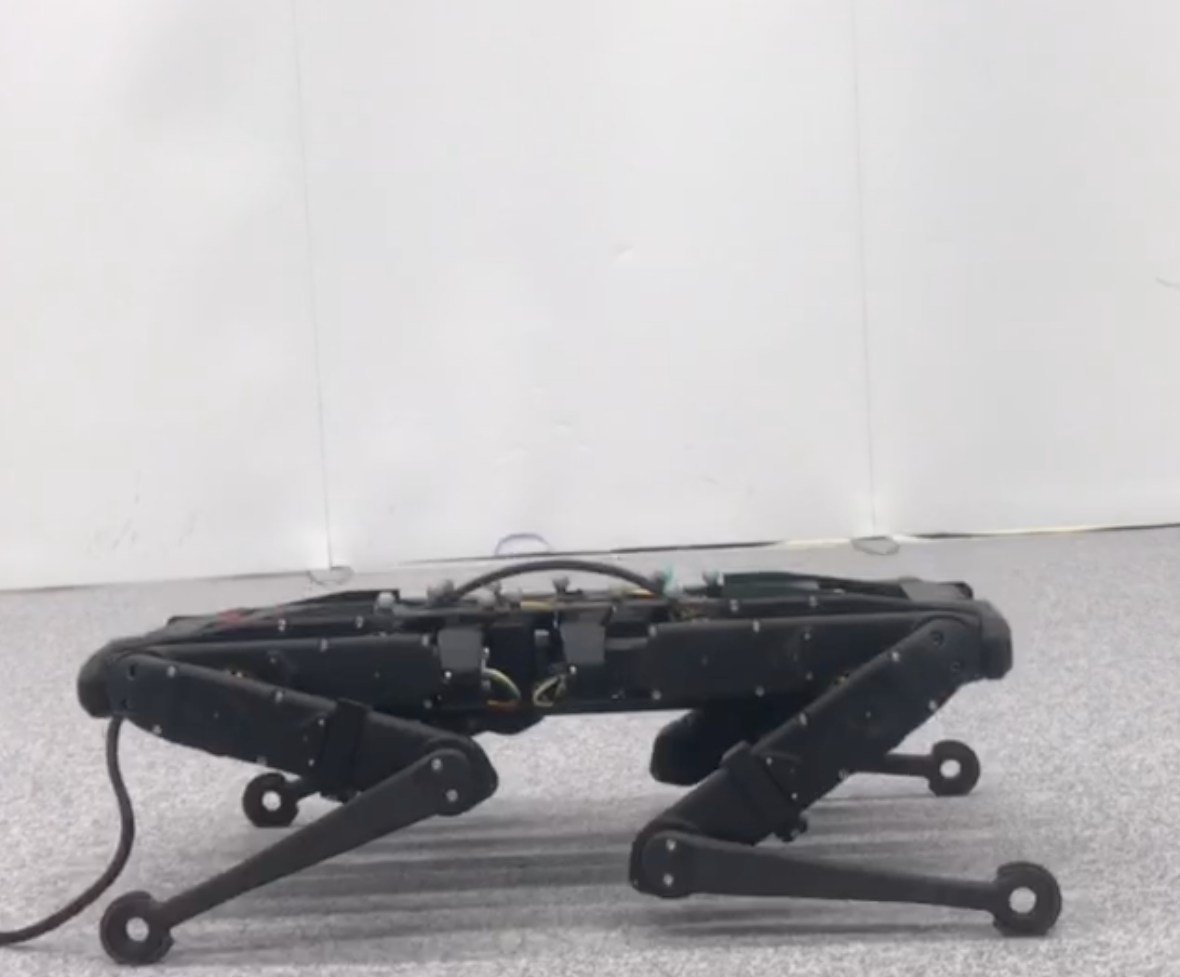}%
	\caption[]{Snapshots of the third motion scenario (jump in place).}
	\label{fig:jump_snapshots}
	\vspace{-4.5mm}
\bigskip
    \scalebox{0.85}{\input{files/jump.pgf}}
    \caption{Motion \#3: Jump in place with contact switching. Estimation of the CoM position $(m)$ (top row), linear momentum $(\frac{kgm}{s})$ (middle row) and angular momentum $(\frac{kgm^2}{s})$ (bottom row); computed centroidal states in \textcolor{blue}{blue}, and EKF in \textcolor{red}{red}.}
    \label{fig:jump}
    \vspace{-5mm}
\end{figure*}
%
\subsection{Motion \#2: Forward Trotting}
In the second scenario, we consider a more complex motion involving intermittent
 contact switching. In this test, the robot moves forward approximately one meter in a trotting gait (Fig. \ref{fig:trot_snapshots}). Contact detection is performed by projecting the joint torques of each leg through the Jacobian into the space of endeffector forces, and applying a force threshold to determine whether each foot is in contact.  As shown in Fig. \ref{fig:trot}, in this case there are large discontinuities in the computed states (blue) due to the contact switches; in practice, these anomalies could destabilize a control loop if these states are used for feedback. However, the proposed torque measurement-based EKF (red) reduces the magnitude of these discontinuities, resulting in state estimates which are more suitable to use in a control loop.
\subsection{Motion \#3: Jump in Place}
The third motion we consider is a jump in place (Fig. \ref{fig:jump_snapshots}). This motion is particularly challenging due to the large impact induced during landing that could be problematic if a direct force/torque sensor at the endeffector is used to estimate the centroidal states as in \cite{rotella2015humanoid}. However, since we are instead using torque measurements in the proposed estimator, we benefit from the fact that Solo12's structural and drive system damping filters out the effect of the impact from measured torque considerably. This helps prevent the estimator from diverging in the presence of large impacts. As demonstrated by the results shown in Fig. \ref{fig:jump}, the proposed estimator is not significantly affected by the impact force during landing, and can thus filter out the measurement noise even in the presence of large impact forces during dynamic maneuvers.
\section{Conclusions and Future Work}
In this paper, we proposed to use joint torque measurements for the centroidal momentum state estimation using an EKF for legged robots. First, we exploited the relationship between the joint torques and centroidal states by projecting the whole-body dynamics of the robot into the nullspace of the contact constraints. Then, we used the resulting dynamics as the process model of an EKF with joint torque input to estimate the centroidal states. One of the key advantages of our approach is that we can estimate the required quantities without relying on Force/Torque sensors as direct measurements. We evaluated the performance of our centroidal momentum estimator for three different gaits on the quadruped robot Solo12. The results of these experiments showed that the estimated states from our approach have considerably less noise when compared to the naive computation of the centroidal states from the measured encoder values and estimated base states while maintaining minimal delay.

In the future, we would like to test the performance of closed-loop momentum control using the estimated centroidal states from the proposed EKF. Furthermore, we would like to mitigate the effects of dynamic model uncertainties by simultaneously estimating the centroidal states and performing online model identification. We are also interested in extending the estimation problem to enable identification of the (non-rigid) contact model parameters, which could be used to adapt the controller when the surface properties are changed.

\bibliography{master} 
\bibliographystyle{ieeetr}

\end{document}